\pdfoutput=1

\documentclass[11pt]{article}

\usepackage[preprint]{acl}

\usepackage{times}
\usepackage{latexsym}

\usepackage[T1]{fontenc}

\usepackage[utf8]{inputenc}

\usepackage{microtype}

\usepackage{inconsolata}
\usepackage{hyperref}
\usepackage{amssymb}
\usepackage[utf8]{inputenc}
\usepackage{CJK}

\usepackage{graphicx}
\usepackage{booktabs}

\title{Multi-Modal Multi-Granularity Tokenizer for Chu Bamboo Slip Scripts}

\author{Yingfa Chen, Chenlong Hu, Cong Feng, Chenyang Song \\ {\bf Shi Yu}, {\bf Xu Han}, {\bf Zhiyuan Liu}, {\bf Maosong Sun}\\
        NLP Group, DCST, IAI, BNRIST, Tsinghua University, Beijing, China \\
        \texttt{yingfa-c24@mails.tsinghua.edu.cn}
}

\begin{document}

\begin{CJK*}{UTF8}{gbsn}

\maketitle
\begin{abstract}
This study presents a multi-modal multi-granularity tokenizer specifically designed for analyzing ancient Chinese scripts, focusing on the Chu bamboo slip (CBS) script used during the Spring and Autumn and Warring States period (771-256 BCE) in Ancient China. Considering the complex hierarchical structure of ancient Chinese scripts, where a single character may be a combination of multiple sub-characters, our tokenizer first adopts character detection to locate character boundaries, and then conducts character recognition at both the character and sub-character levels. Moreover, to support the academic community, we have also assembled the first large-scale dataset of CBSs with over 100K annotated character image scans. On the part-of-speech tagging task built on our dataset, using our tokenizer gives a 5.5\% relative improvement in F1-score compared to mainstream sub-word tokenizers. Our work not only aids in further investigations of the specific script but also has the potential to advance research on other forms of ancient Chinese scripts.\footnote{The code and data are available at: \url{https://www.github.com/THUNLP/Chujian}} 

\end{abstract}


\section{Introduction}

Deep neural networks have demonstrated remarkable success in various natural language processing tasks \citep{gpt4,Touvron2023LLaMAOA} as well as the analyses of ancient languages \citep{sommerschield2023machine}. Inspired by these former works, we aim to apply deep learning to the analysis of ancient Chinese scripts. However, this application faces three challenges: (1) Most of these ancient scripts are stored as images, which are more difficult to analyze than texts. (2) A large proportion of the characters is rare or undeciphered, making it challenging to train data-driven neural networks. This also implies that the widely-used subword tokenizers such as BPE \citep{sennrich-etal-2016-neural} and SentencePiece \citep{Kudo2018SentencePieceAS} fall short because the neural networks struggle to learn informative representations of the rare and undeciphered characters. (3) Current tokenizers struggle to generalize to unseen materials, in which there is a considerable ratio of out-of-vocabulary (OOV) characters.

\begin{figure}[!t]
    \centering
    \includegraphics[width=0.99\linewidth]{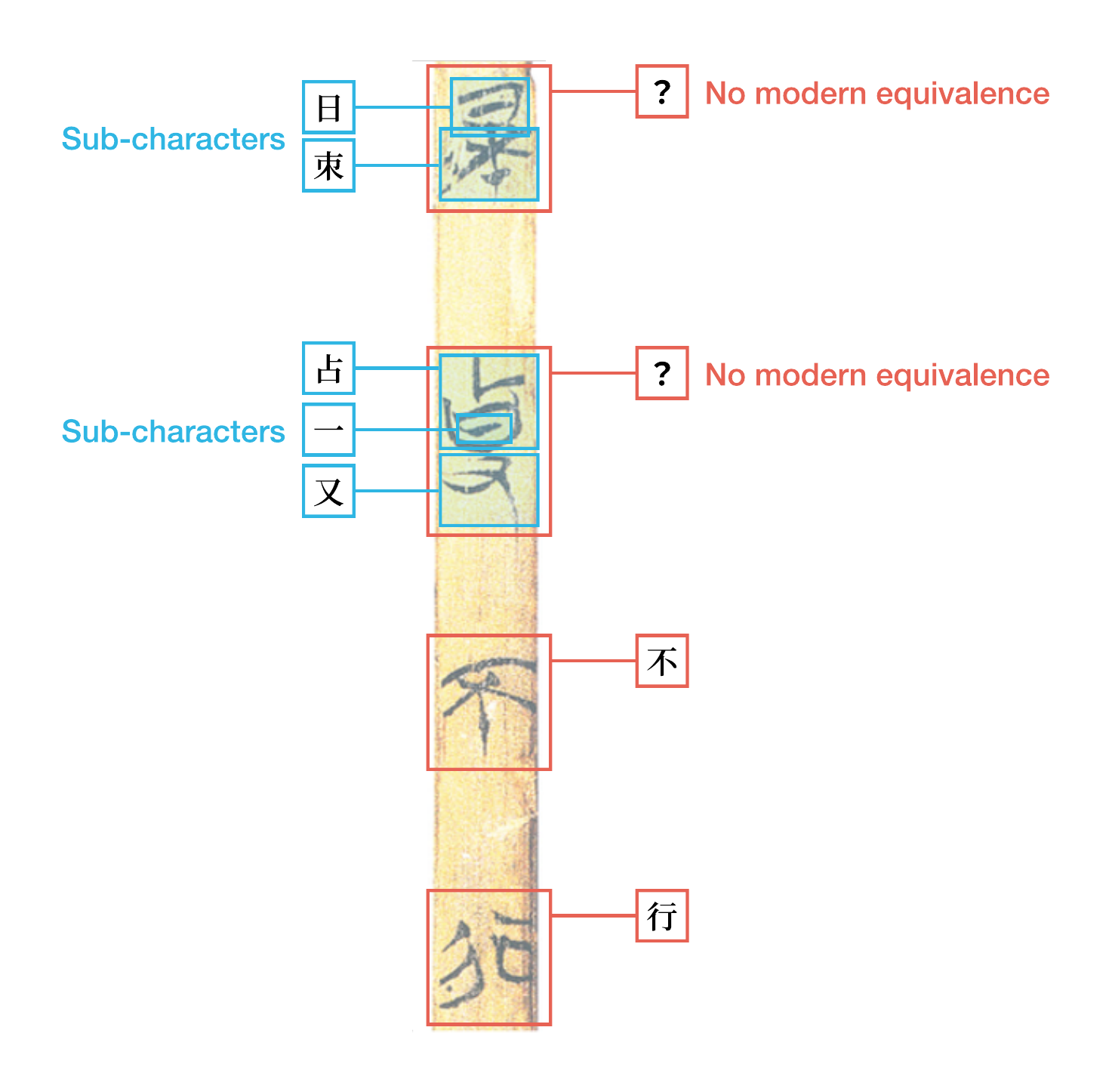}
    \caption{Overview of our proposed tokenizer on an example. Each ancient character is mapped to a modern character if possible. Otherwise, the tokenizer rolls back to decomposing the character into sub-character units, potentially containing useful information. One possible deciphering of the text is ``At first, action is not simple''. The slip shown is the 14th slip in Zhonggong document from the Shanghai Museum Slips.}
    \label{fig:method-overview}
\end{figure}

To overcome these challenges, we propose a novel multi-modal multi-granularity tokenizer tailored for ancient Chinese scripts, focusing on the 2000-year-old Chu bamboo slip (CBS) script from ancient China. The tokenization pipeline begins by detecting and ordering the characters in image scans of the raw materials into a sequence of character images. Next, each character is recognized within a pre-defined vocabulary. If the recognition confidence is low, the tokenizer rolls back to tokenizing the character into \textit{sub-character components} (components that make up Chinese characters and are larger than a stroke, and smaller than a character) which may contain rich information about the semantics or phonetics of the text \cite{radical-enhanced-chinese-character-embedding, subchar-japanese,si-etal-2023-sub}.


To demonstrate the effectiveness of our tokenizer, we collect and release the first dataset of CBS texts. It contains 102,722 annotated CBS character images, from 5,033 slips and 164 documents. To facilitate further investigation, we have developed a user-friendly platform where researchers with different expertise can access and analyze the dataset with ease. 
The proposed tokenizer significantly outperforms the existing baselines, especially on the task of part-of-speech tagging.

The main contributions of this study can be summarized as follows:
\begin{enumerate}
    \item We collect, process, and release CHUBS, the first large-scale dataset on Chu Bamboo Slip script in a format that is convenient for typical NLP workflows.
    \item We propose an annotation scheme for provided useful information about the sub-character features of CBS scripts to address the large proportion of out-of-vocabulary characters prevalent in CBS.
    \item Based on the sub-character annotations, we propose a multi-granularity tokenizer that outperform ordinary character-based tokenizers on downstream tasks.
    \item We build a platform for easy access to the data for researchers of all background to facilitate future research.
\end{enumerate}

\section{Related Works}

\paragraph{Tokenization}

Tokenization is the process of splitting a sentence into units. It is essential to current natural language processing techniques and have an integral impact on downstream performance \citep{Mielke2021BetweenWA}. Current NLP tokenizers accept text sequences as inputs and split them into pieces that are then turned into integers to be handled by neural networks. In this work, although the tokenization process start from the image scan of text inscriptions, the goal is to convert the raw representation into a sequence of simple representations that are easy for the pipeline to handle. Therefore, we call our method a tokenization pipeline. 

\paragraph{Chinese Tokenization}

Regarding Chinese characters, most existing tokenization methods operate on the character level. Each token is either once character or a combination of character \citep{si-etal-2023-sub}. Such method disregard the fact that each Chinese character is composed of components that encode information that may be useful for analyzing the language. Numerous works have shown that tokenizing characters at the sub-character level can improve the downstream performance of Chinese, Japanese, and Korean neural models. Some notable works include \citet{radical-enhanced-chinese-character-embedding,component-enhanced-chinese,Song2018JointLE,si-etal-2023-sub}, which have shown that utilizing sub-character components can improve the quality of learned embeddings, as measured by improved performance or efficiency in a wide range of language understanding tasks compared to conventional tokenizers. For pre-trained language models, \citet{si-etal-2023-sub} show that converting Chinese characters to sub-character sequences can improve the efficiency and robustness in general language understanding. In language generation, ~\citet{Wang2022BreakingTR} have showed that using stroke information can improve English-Chinese translations.

\paragraph{Deep Learning Applications in Ancient Scripts}

As a result of the recent advances in the capabilities of deep neural networks in computer vision and natural language processing, there have been numerous works that utilize deep learning methods to assist research in ancient scripts \citep{sommerschield2023machine}. Some examples include ancient Greek \citep{assael2022restoring}, Devanagari \citep{narang2021deepnetdevanagari}, ancient Chinese \citep{zhang2021feature}, ancient Japanese \citep{clanuwat2019kuronet}, etc. To the best of our knowledge, our work represent the first attempt to apply deep learning methods in the processing of Chu bamboo slips.

\section{Dataset}

We begin with a brief introduction to the background of the CBSs (Section~\ref{sec:backgroud}). Then, we describe the collection process of our dataset called CHUBS (\textbf{CHU} \textbf{B}amboo \textbf{S}lips) (Section~\ref{sec:dataset}). Finally, we present an open platform for convenient access to our data, especially for researchers of different backgrounds (Section \ref{sec:platform}).

\subsection{Chu Bamboo Slips} \label{sec:backgroud}

\begin{table*}[!t]
    \centering
    \begin{tabular}{l|cccc}
        \toprule
        \textbf{Source name} 
            & \textbf{Chinese name}  
            & \textbf{\# documents}         
            & \textbf{\# slips} 
            & \textbf{\# characters} \\
        \midrule
        Tsinghua University Slips & 清华简   & 50  & 1,402 & 31,468 \\
        Shanghai Museum Slips    & 上博简   & 60  & 881 & 25,795 \\
        Baoshan Slips       & 包山简   & 4   & 337 & 12,647 \\
        Guodian Slips       & 郭店简   & 18  & 705 & 11,865 \\
        Geling Slips        & 葛陵简   & 8   & 743 & 6,209 \\
        Zenghouyi Slips     & 曾侯乙简 & 4   & 198 & 6,016 \\
        Jiudian Slips       & 九店简   & 2   & 232 & 2,956 \\
        Wangshan Slips      & 望山简   & 3   & 273 & 2,218 \\
        Changtaiguan Slips  & 长台关简 & 3   & 148 & 1,504 \\
        Zidanku Silk        & 子弹库帛 & 7   & 7   & 1,471 \\
        Yangtianhu Slips    & 仰天湖简 & 1   & 42  &  335 \\
        Wulipai Slips       & 五里牌简 & 1   & 18  &  109 \\
        Xiyangpo Slips      & 夕阳坡简 & 1   & 2   &   54 \\
        Ynagjiawan Slips    & 杨家湾简 & 1   & 38  &   41 \\
        Caojiagang Slips    & 曹家岗简 & 1   & 7   &   34 \\
        \midrule
        Total               &         & 164 & 5,033  & 102,722 \\
        \bottomrule
    \end{tabular}
    \caption{The amount of data from different sources of our collection of CBSs.}
    \label{tab:data_source-detailed}
\end{table*}

CBSs are the writing materials used in ancient China during the Warring States period over two thousand years ago, and the earliest known large-scale form of calligraphic writing\footnote{Some Oracle Bone Script were formed by brushes, but only in extremely small amounts.}. The study of it holds great linguistic, historical, and cultural value, especially for East Asian scripts.
The content includes, for instance, the oldest known records of ancient classics such as the \textit{Book of Documents} (also the \textit{Classic of History}, Chinese: 尚书) and \textit{Classic of Poetry} (Shijing, Chinese: 诗经).

The slips survived over two thousand years mainly because when they were submerged in water until excavation, protecting them from oxidation. For the same reason, most current slips are found along the Yangtze River. As shown in Figure \ref{fig:example_cbs}, the form of the slips is highly regular, most are 45cm long and 0.6cm wide. The longer slips typically carry between 27 to 38 characters. Multiple slips are tied together to form documents. A real example of a CBS is given in Appendix \ref{sec:appendix-cbs-example}.

\subsection{CHUBS} 
\label{sec:dataset}

Digitizing and understanding CBSs, especially in the view of natural language processing, are of great value in promoting history, culture, and art research studies. 
However, to the best of our knowledge, there is no public large-scale collection of CBS dataset prepared in an accessible format that is convenient for usage in typical workflows within the machine learning community. 
Thus, to facilitate the application of machine learning to aid research in CBS, we collect and publish the first dataset of CBS inscriptions, called CHUBS. It includes high-quality scanned images of the slips and their text annotations.

\subsubsection{Data Source}

All data is extracted and processed from publicly released textbooks or records by paleographers, containing image scans and transcriptions of a set of bamboo slips from certain excavation projects. We supplemented the materials with some missing transcriptions and extracted the images of the characters from the slip images.

These materials are widely known in the community of paleographers in ancient Chinese scripts. Our contribution is that we are the first to compile these materials into an easily accessible format for the application of machine learning methods. We have been careful to ensure that there is no restriction on the use of these materials, and the data will be released under a permissive license.

Since all image scans are extracted and processed from publicly released textbooks containing unearthed materials from various sources and different periods, variations between scans produced by different teams are inevitable. For example, some scans are black, while others are in color.

Table \ref{tab:data_source-detailed} lists each of the data sources as well as the number of documents, slips, and characters from each source. 
It is worth noting that many of the sources do not have an official English name. Therefore, we only give the pinyin transcription of the Chinese name. We highly suggest interested readers use the Chinese name when possible for future research.


\subsubsection{Annotating Sub-Character Components}
\label{sec:annotating-sub-character-components}

Each character is annotated with modern Chinese text.
However, manual inspection reveals that at least 27\% of the characters in our dataset are not within the set of modern Chinese words\footnote{A word may consist of multiple characters.} (these characters do not have a UTF-8 encoding). In other words, 27\% of the detected characters are out of vocabulary (OOV) if we tokenize them on character-level granularity. The upper two characters in Figure \ref{fig:method-overview} are examples of such OOV characters.

There are two reasons for this high proportion of OOV CBS characters:
\begin{enumerate}
    \item The CBS character has not yet been deciphered due to drastic changes in character forms or material degradation.
    \item The CBS character does not have a modern Chinese equivalent (but experts believe that they know the meaning of the character). 
\end{enumerate}
Such CBS characters are annotated with a set of \textit{sub-character components} such as radicals or \textit{pianpang}\footnote{Components of Chinese characters traditionally used for indexing in dictionaries.}. For instance, assuming the character ``想'' (pronunciation: \textit{xiang}) does not have a modern equivalence, it may be labeled as ``相心''\footnote{We have refrained from using a more advanced encoding system (such as including the positioning of the components) to keep the annotation cost low.} (pronunciation: \textit{xiang xin}).
If even such sub-character components are unrecognizable, it is annotated with a placeholder to indicate that the character is unrecognizable.

However, there is no common consensus on how to split Chinese characters into sub-character components. Our approach is based on the philosophy that each unit shoudl be semantically or phonetically meaningful (i.e., it is a morpheme or a phoneme). This is because we hypothesize that further splitting such units does not provide additional useful information about the text, but may introduce noise or result in unnecessarily lengthy token sequences.

Concretely, we request an expert in the field (with a Ph.D. degree studying CBS) to annotate each CBS character with the corresponding sub-character components. One possibility is to label the pianpang. However, this has two main limitations when applied to CBS scripts. Firstly, CBS characters are very different from modern Chinese and not every CBS character has a pianpang. Secondly, we want to retain as much information about the character as possible, so we need a method for annotating the semantics or phonetics of the part of the characters that is not the pianpang as well.

\subsubsection{Sub-Character Component Annotation Scheme}

Addressing the above limitations, our final annotation procedure is as follows. For a given CBS character, if it is already labeled with a modern Chinese character (i.e., it is not OOV), we keep it as it is. Otherwise, we first identify it as one of the three types of Chinese characters: \textbf{logograms} (\textit{xiangxing} characters, Chinese: 象形字), \textbf{semantic-phonetic compound characters} (\textit{xingsheng} characters, Chinese: 形声字), and \textbf{phonograms} (\textit{jiajie} characters, Chinese: 假借字). Such classification of Chinese characters was first introduced by \cite{mengjia}, and is commonly taught in Chinese schools\footnote{This categorization scheme is called ``three category theory'' (\textit{san shu shuo}, Chinese: 三书说), but there are also other categorization methods. Two notable instances are ``four category theory'' and ``six category theory''.}. 
Then, we start with a sub-character vocabulary with 540 items introduced by \textit{Shuowen Jiezi} \cite{xushen1963}, a well-known Chinese dictionary released around 100 CE during the Eastern Han dynasty.
\begin{itemize}
    \item For \textbf{semantic-phonetic compound characters}, we split them into the semantic and phonetic parts (the former is always a logogram), and apply the following rules.
    \item For \textbf{logograms} and \textbf{phonograms}: we try to split it into components of the current sub-character vocabulary. If there exists a part of the character that is not and does not include any of the current sub-character components, we add that part as a sub-character component into the vocabulary.
\end{itemize}
Repeating this process for all characters in our library results in 798 sub-character components in total, which makes up our final sub-character vocabulary.

We emphasize that the vocabulary construction may have considerable impact on the downstream performance, but it is out of the scope of this thesis work.


\subsection{Open Platform}
\label{sec:platform}

To better foster future research in CBS scripts, we build and release a platform to make accessing our data more convenient for researchers from different backgrounds. The platform allows the download of the entire collection as well as searching particular images based on the text annotation, origin, and character appearance (searching by hand-written strokes), which is essential for searching for characters without modern Chinese equivalents.
Further, this platform also features pipeline processing capabilities for CBS, including detecting, recognizing, and retrieving characters, significantly reducing both time and human resources for experts. Specifically, for a CBS image, it can detect each character and recognize it with our multi-modal tokenizer. Appendix \ref{sec:appendix-platform} displays a screenshot of this platform.

\section{Multi-Modal Multi-Granularity Tokenizer}

In summary, our tokenizer consists of multiple neural networks that perform object detection and classification in a pipeline. The input is the image of the material containing the Ancient inscriptions. The pipeline consists of the following steps:
\begin{enumerate}
    \item The characters in the bamboo slip are detected using an object detection model, cropped then ordered into a sequence based on their location.
    \item Each image is fed to a character recognition that maps the CBS characters into a modern Chinese character/word.
    \item If the classification confidence is lower than a certain threshold, the tokenizer falls back to sub-character analysis by recognizing the sub-character components of the character. 
\end{enumerate}
The output is a sequence where each element is either a single character or a set of sub-character components. The classification confidence threshold is typically determined using a validation set of examples from the downstream task.

\subsection{Sub-Character Recognition}

As mentioned in Section \ref{sec:annotating-sub-character-components}, many characters in our dataset are not within the set of modern Chinese words. For such characters, assigning a unique class would not be conducive, because the class label may not help us better understand the ancient text. Therefore, we propose to recognize the sub-character components\footnote{We use ``components'' to refer to any consistent and frequent set of strokes smaller than or equal to a character.} of the characters instead. This may be beneficial for downstream tasks because Chinese character components may represent rich information about the phonetics and semantics of the character.

This is done with a multi-label classifier whose vocabulary is simply the set of 798 sub-character components we have annotated in CHUBS.




\section{Experimental Details}

\subsection{Models}

\paragraph{Character Detection}

Specifically, we employ the YOLOv5 model~\citep{yolov5}, one of the most used versions in the YOLO series \citep{redmon2016you}.
We train this model on the CBS images annotated by domain experts.

\paragraph{Character and Sub-Character Recognition}

For both character and sub-character recognition, we try both ResNet \citep{resnet} and Visual Transformer (ViT) \citep{vit}, which are two strong models with great capabilities in image classification. We use roughly the same number of parameters for both architectures. The difference between character and sub-character recognition is the number of classes and that the former is an ordinary multi-class classification while the latter is a multi-label classification.

Specifically, we start from commonly used public checkpoints, the official \texttt{resnet152} model of PyTorch and the ViT by \citet{visual-transformers}\footnote{\url{https://huggingface.co/google/vit-base-patch16-224}}. These model checkpoints are pre-trained on ImageNet \citep{imagenet}, and we finetune them on CHUBS.

\subsection{Training Data}

\paragraph{Detector Training Data}

To train the CBS character detector, an expert paleographer is asked to manually annotate a small number of CBS. 
The annotations are then quality-checked by other authors. In total, 177 image scans of bamboo slips from Tsinghua University Slips were annotated, of which 141 were used as training data, and 36 for validation. We emphasize that this annotation process is rather simple because most CBS characters are very easy to identify in the image scans.

\paragraph{Classifier Training Data}

The character and sub-character recognizer are simply trained on CHUBS, since the data already contains all supervision needed.  
The frequency distribution of the characters follows a Zipfian distribution, so approximately half of the characters only appear once in the dataset. To ensure that each class contains enough data for both training and testing, we discard characters that have less than $k$ images (we use $k=3, 10$ in character recognition and $k=2, 20$ in sub-character recognition). We then split the data into training, validation, and test sets by an 8:1:1 ratio, while ensuring that the test set has at least one example from every class.

\subsection{Training Details}

All training experiments are conducted on an A100 GPU, and implemented with PyTorch. We use the Adam optimizer \citep{kingma2020method} and a learning rate scheduler that decays by 0.9 after every epoch. We only search different batch sizes and maximum learning rates during the hyperparameter search to keep the computational cost low.











\section{Results}

Since the tokenization pipeline has three steps, we first show the empirical performance of each part. Then, we apply the tokenizer on an example downstream task, part-of-speech (POS) tagging, to demonstrate its effectiveness over character-based tokenizers (one CBS character per token). 

\subsection{Character Detection}

The performance of the character detector is shown in Table \ref{tab:result_yolov5}. 
The \textit{near-perfect} F1-score implies that the model is well-suited and robust for CBS characters and that it introduces minimal noise to our tokenization pipeline. Based on these detection results, we then conduct character recognition.

\begin{table}[!ht]
    \centering
    \begin{tabular}{ccc}
        \toprule
        \textbf{Precision}   &   \textbf{Recall} &   \textbf{F1} \\
        \midrule
        0.998    & 0.996 & 0.997 \\
        \bottomrule
    \end{tabular}
    \caption{Character Detection Results with YOLOv5.}
    \label{tab:result_yolov5}
\end{table}

\subsection{Character Recognition}




The result of the character recognizer on the test set is shown in Table \ref{tab:recognition}, in which we can see that ViT consistently outperforms ResNet, which is consistent with the results by the authors of ViT.
The high accuracy indicates that the application of such deep learning offers great practical value.

\begin{table}[!t]
    \centering
    \begin{tabular}{l|cccc}
        \toprule
        \textbf{Model}  
            & \textbf{Top-1} 
            & \textbf{Top-3} 
            & \textbf{Top-5} 
            & \textbf{Top-10} \\
        \midrule
        \multicolumn{5}{c}{$k=3$} \\
        \midrule
        ResNet  & 61.23 & 65.48 & 70.84 & 72.33 \\
        ViT     & 73.48 & 84.65 & 87.45 & 89.95  \\
        \midrule
        \multicolumn{5}{c}{$k=10$} \\
        \midrule
        ResNet  & 72.60 & 83.70 & 87.18 & 90.57 \\
        ViT     & 90.11 & 95.03 & 96.06 & 97.16 \\
        \bottomrule
    \end{tabular}
    \caption{Accuracy (in \%) of character recognition models on the test set. $k$ indicates the minimum occurrence of a character in the dataset.}
    \label{tab:recognition}
\end{table}

\subsection{Sub-Character Recognition}

\begin{table}[!t]
    \centering
    \begin{tabular}{l|ccc}
        \toprule
        \textbf{Method}  & \textbf{Recall} & \textbf{Precision} & \textbf{F1} \\
        \midrule
        \multicolumn{4}{c}{$k=2$} \\
        \midrule
        ResNet  & 84.79 & 77.32 & 80.88 \\
        ViT     & 22.48 & 26.31 & 24.24 \\
        \midrule
        \multicolumn{4}{c}{$k=20$}\\
        \midrule
        ResNet  & 85.70 & 78.31 & 80.19 \\
        ViT     & 28.57 & 28.23 & 28.40 \\
        \bottomrule
    \end{tabular}
    \caption{Recognition result (in \%) of sub-character components of our model.} 
    \label{tab:sub-char-result}
\end{table}

Table~\ref{tab:sub-char-result} shows the performance of the sub-character recognition module. Perhaps surprisingly, ResNet beats ViT by a large margin, which differs from the observation in the character recognition experiments. One possible explanation for this is that each head in the multi-head attention module is responsible for recognizing a certain set of components (or their corresponding features), but the number of classes is too great for the architecture. Further investigations are outside this work's scope.

\subsection{Downstream Task: Part-of-Speech Tagging}
\label{sec:post}

To demonstrate the effectiveness of our multi-granularity tokenizer, we apply it to a part-of-speech (POS) tagging task in the CBS script. 

We create a POS tagging dataset for CBS by manually annotating 1,109 randomly sampled sentences using the BIO (Beginning, Inside, and Outside) format \citep{ramshaw1999text}. This annotation is conducted by an expert in CBS scripts.
Then, we apply our multi-granularity tokenizer and a character-based tokenizer (each character is one token).

Our annotations include the following ten part-of-speeches that are commonly found and analyzed in ancient Chinese:

\begin{enumerate}
    \item Noun (Chinese: 名词, \textit{mingci})
    \item Verb (Chinese: 动词, \textit{dongci})
    \item Conjunction (Chinese: 连词, \textit{lianci})
    \item Adjective (Chinese: 形容词, \textit{xingrongci})
    \item Adverb (Chinese: 副词, \textit{fuci})
    \item Numeral (Chinese: 数量词, \textit{shuliangci})
    \item Modal Particle (Chinese: 语气词, \textit{yuqici})
    \item Pronoun (Chinese: 代词, \textit{daici})
    \item Preposition (Chinese: 介词, \textit{jieci})
    \item Auxiliary Word (Chinese: 助词, \textit{zhuci})
\end{enumerate}

This dataset will be publicly released along side with our CHUBS dataset and training code.

When splitting characters into sub-character components, the label corresponding to the components is the same as the label for the original character. Then, a special token representing the boundary between each character is added to the sides of the sequence of components for each character. The predictions for these special tokens are ignored.

For the downstream model, we tune a large language model for this task using in-context learning. Specifically, we randomly sample 10 examples from the training data to use as in-context demonstrations and prompt the LLM to generate the predicted entities and the types as a Markdown list. The actual prompt template will be given along with the code after the review period. We use GPT-3-Turbo with default hyperparameters and repeat the experiments with 10 random seeds to ensure reproducibility.


The result is shown in Table \ref{tab:result-downstream}. We observe that using our multi-granularity tokenizer can significantly (with a p-value of $0.0079$ in a t-test) improve the POS tagging performance of the downstream model, as we have expected.

\begin{table}[!t]
    \centering
    \begin{tabular}{l|ccc}
        \toprule
        \textbf{Tokenizer} & \textbf{Recall} & \textbf{Prec.} & \textbf{F1} \\
        \midrule
        Character-based             & 47.9 & 43.8 & 45.3 \\
        Multi-granularity    & 50.2 & 46.1 & 47.8 \\ 
        \bottomrule
    \end{tabular}
    \caption{The part-of-speech performance (in \%) when using a conventional character-level tokenizer (Char-Tokenizer) and our multi-granularity tokenizer.}
    \label{tab:result-downstream}
\end{table}

\section{Conclusion and Discussions}

We have proposed a multi-modal multi-granularity tokenizer for better analyzing ancient Chinese scripts than the existing more popular sub-word tokenizers. We have also collected the first large-scale multi-modal dataset of CBS text with an open platform targeted at audiences of different backgrounds. We believe this work is an important step in leveraging deep learning methods in the research of East Asian scripts.


\paragraph{What are the differences between multi-granularity tokenizers and mainstream tokenizers?}

Currently, most tokenizers are a kind of ``subword tokenizer''. This includes Byte-Pair Encoding (BPE) \citep{sennrich-etal-2016-neural} and SentencePiece \citep{Kudo2018SentencePieceAS}, used in GPT-4 \citep{gpt4} and LLaMa \citep{Touvron2023LLaMAOA}, respectively. The tokens in these tokenizers are often called \textit{sub-words} (sequences of characters that are smaller than space-delimited words but larger than letters). For Chinese, mainstream tokenizers usually treat each Chinese character as an atomic unit. In contrast, our multi-granularity tokenizer splits each Chinese character into smaller sub-character components and provides this information to the downstream neural network.

\paragraph{Why is identifying sub-character components conducive to downstream tasks?}

Tokenizers that treat each character as an atomic unit generally work for phonetic languages such as Indo-European languages because splitting phonetic letters into smaller components typically provides little to no additional information\footnote{One possible example task that we hypothesize might benefit from splitting Latin characters into multiple tokens is for answering questions about the shape of the letters.}. 
However, for ideographic languages such as Chinese, components of a character may encode rich information about the semantics or phonetics of the characters. For common characters, the model may be able to learn such information automatically from that distribution of co-occurring characters.
However, for infrequent characters or unknown characters, the sample size of co-occurring characters is too small. Although some works have shown that language models can implicitly learn the letter composition (which is a kind of sub-token information) of tokens \citep{sub-character-tokenization,hiraoka2024knowledge}, it is reasonable to hypothesize that such information requires large amounts of training data and tokenization at the sub-character level can provide conductive bias that either enhances performance or reduces the amount of data required to achieve the same performance.

\section*{Limitations}

In terms of the performance of the tokenizer, there are many possible methods for improving the effectiveness of the components of our tokenizer, such as pre-training on a corpus of modern text, larger/better model architectures, and better data pre- or post-processing. Moreover, augmenting the tokenizer with more knowledge about the history may help. But, we have not employed more tricks to keep our analysis simple.

Also, although we have demonstrated the effectiveness of our tokenizer on the CBS script, it may be less effective on other scripts due to variations between scripts. However, since many other scripts face the same challenges highlighted in the introduction, our method should still have a performance advantage over conventional tokenizers. Additionally, due to the annotation cost, we have only investigated our tokenizer's effectiveness on one downstream task.

\section*{Ethical Concerns}

This work presents a new dataset on the Chu bamboo slips, a writing material from ancient China over two thousand years ago. We also introduce a new tokenizer for better processing ancient Chinese scripts with a large number of characters that do not have modern Chinese correspondence. The goal is to advance research in this ancient script as well as other forms of ancient Chinese scripts, which should not have significant ethical implications. However, the original content from these raw materials may have ethical implications for certain groups, but since these are existing historic materials, we do not make efforts to censor any content.

\section*{Acknowledgements}

This work is supported by the National Key R\&D Program of China (No. 2022ZD0116312), National Natural Science Foundation of China (No. 62236004), and Institute Guo Qiang at Tsinghua University.

\newpage

\bibliography{custom,anthology}

\appendix

\newpage

\section{An example of a CBS Material}
\label{sec:appendix-cbs-example}

\begin{figure}[!t]
    \centering
    \includegraphics[height=0.88\textheight]{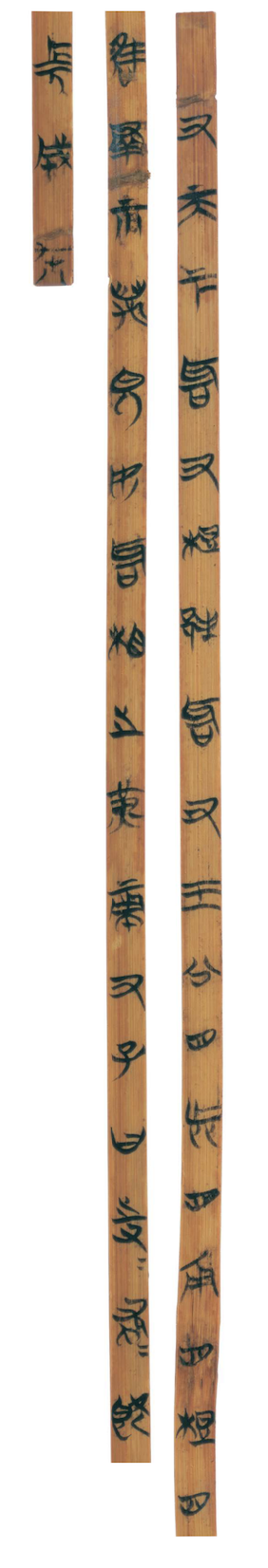}
    \caption{An example of a CBS material. The slip shown is the 98th slip of the ``Wu Ji'' from Tsinghua University Slips.}
    \label{fig:example_cbs}

\end{figure}

To better understand the nature of CBS, we give an example of a CBS material in our dataset in Figure \ref{fig:example_cbs}. The text is read from top to bottom and right to left.

\section{Open Platform}
\label{sec:appendix-platform}

\begin{figure*}
    \centering
    \includegraphics[width=0.99\linewidth]{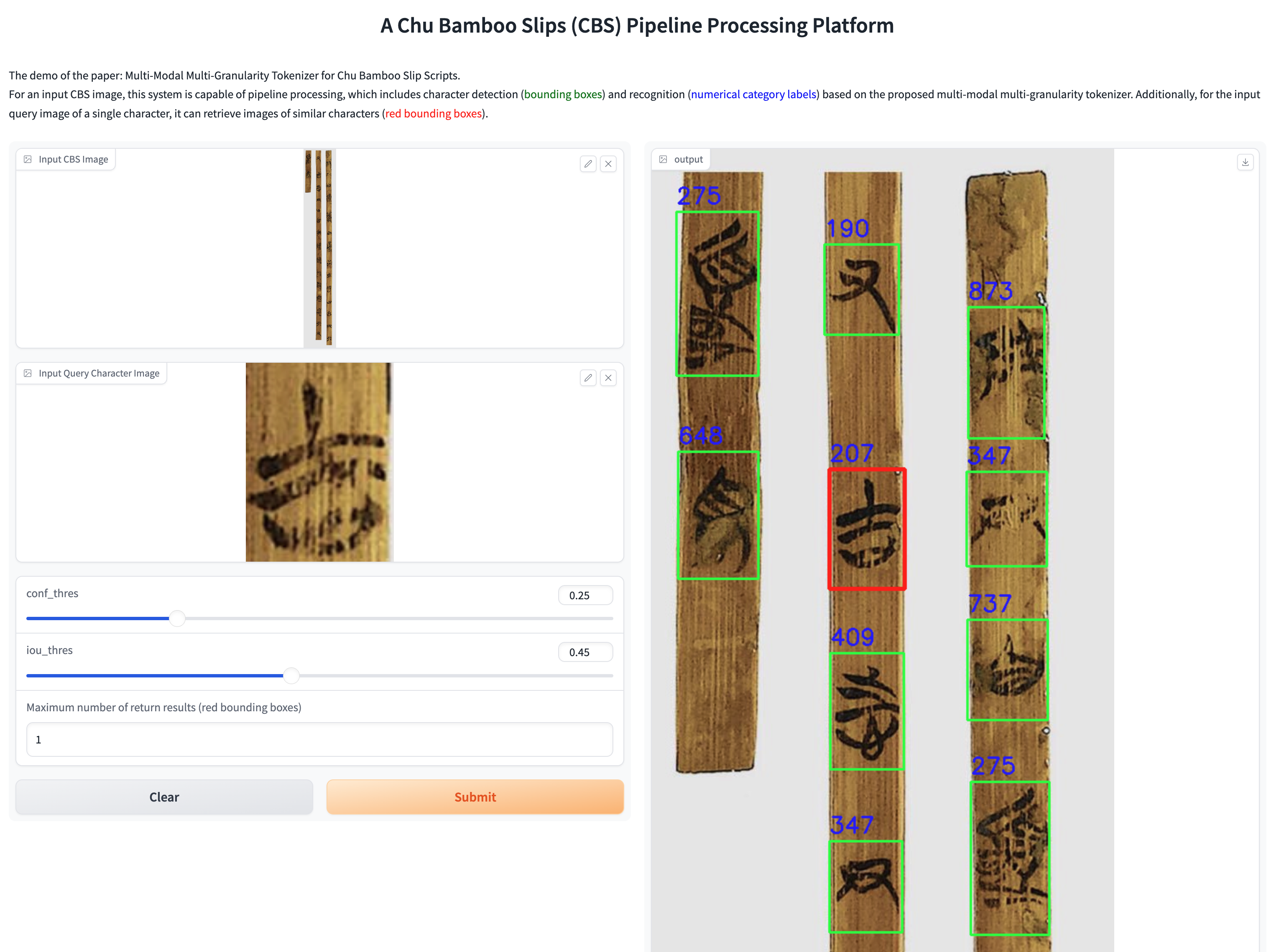}
    \caption{A screenshot of our platform for accessing the dataset and a demo of our tokenizer.}
    \label{fig:platform-screenshot}
\end{figure*}

The platform described in Section \ref{sec:platform} will be launched after the anonymous review process. A screenshot of it is shown in Figure \ref{fig:platform-screenshot}. The platform is a website, and the interaction system was implemented using the Gradio library.

\section{AI-Assistant-Related Statement}

AI-assisted tools were used for error-checking in writing this paper, and for code-completion during the implementation of the experiments.

\end{CJK*}

\end{document}